\documentclass[10pt,letterpaper,twocolumn]{article}

\usepackage[T1]{fontenc}
\usepackage[utf8]{inputenc}
\usepackage[margin=0.75in]{geometry}
\ifx\pdfminorversion\undefined\else\pdfminorversion=7\fi
\usepackage{mathptmx}
\usepackage{hyperref}
\usepackage{url}
\usepackage{booktabs}
\usepackage{amsmath,amssymb,amsfonts}
\usepackage{array}
\usepackage{makecell}
\usepackage{multirow}
\usepackage{tabularx}
\usepackage{graphicx}
\usepackage[table]{xcolor}
\usepackage{cite}
\usepackage{dblfloatfix}
\usepackage{placeins}
\usepackage{pifont}
\usepackage[font=small,labelfont=bf]{caption}

\newcommand{\cmark}{\ding{51}}
\newcommand{\xmark}{\ding{55}}

\makeatletter
\providecommand{\insert@pcolumn}{\insert@column}
\makeatother

\graphicspath{{figures/}}

\definecolor{mvrowgray}{gray}{0.90}
\definecolor{mvgain}{RGB}{0,128,70}
\definecolor{mvloss}{RGB}{190,42,45}
\definecolor{mvneutral}{gray}{0.35}

\newcommand{\pending}{\textemdash}
\newcommand{\best}[1]{\textbf{#1}}
\newcommand{\second}[1]{\underline{#1}}
\newcommand{\abest}[1]{\textbf{#1}}
\newcommand{\asecond}[1]{\underline{#1}}
\newcommand{\gain}[1]{\,{\scriptsize\textcolor{mvgain}{(#1)}}}
\newcommand{\loss}[1]{\,{\scriptsize\textcolor{mvloss}{(#1)}}}
\newcommand{\same}[1]{\,{\scriptsize\textcolor{mvneutral}{(#1)}}}
\newcommand{\oursbar}[9]{\rowcolor{mvrowgray}#1 & #2 & #3 & #4 & #5 & #6 & #7 & #8 & #9}
\newcommand{\realoursbar}[5]{\rowcolor{mvrowgray}#1 & #2 & #3 & #4 & #5}
\newcommand{\appsection}[2]{%
  \refstepcounter{section}%
  \par\medskip
  \noindent{\large\bfseries \Alph{section}.~#2}\label{#1}\par\smallskip
}

\title{MV-Actor: Aligning Multi-View Semantics and Spatial Awareness\\for Bimanual Manipulation}

\author{
  \normalsize
  Yinchen~Tian\textsuperscript{1}\quad
  Huan~Li\textsuperscript{2}\quad
  Muyao~Peng\textsuperscript{1}\quad
  Xi~Wang\textsuperscript{3}\quad
  Yan~Wang\textsuperscript{2}\quad
  You~Yang\textsuperscript{1,*}\\[4pt]
  {\small
  \begin{tabular}{c}
  \textsuperscript{1}School of Electronic Information and Communications,\\
  Huazhong University of Science and Technology, Wuhan, China\\
  \textsuperscript{2}Institute for AI Industry Research (AIR), Tsinghua University, Beijing, China\\
  \textsuperscript{3}AIR Wuxi Innovation Center, Tsinghua University, Wuxi, China\\
  \textsuperscript{*}Corresponding author
  \end{tabular}}
}
\date{}

\begin{document}
\twocolumn[
\maketitle
\vspace{-1.5em}
]
\begin{center}
\textbf{Abstract}
\end{center}
\noindent Robotic manipulation has been widely applied in industrial scenarios. Compared with single-arm manipulation, bimanual manipulation is equipped with multiple cameras to capture information from different viewpoints. However, existing multi-view policies encode each view independently or fuse view features shallowly, resulting in limited sharing semantic perception and unreliable spatial awareness. In this paper, we propose \textbf{MV-Actor}, a multi-view perception framework that builds a unified semantic-spatial representation for bimanual manipulation. First, MV-Actor performs Multi-view Semantic Interaction to share semantic perception across views. Then it uses Semantic-Spatial Token Interaction to ground visual semantics with feed-forward reconstruction model features and acquire reliable spatial awareness. Finally, a Guided Metric Depth Repair module refines degraded sensor depth to provide more reliable metric anchors under consumer-grade depth noise. In simulation experiments conducted on the PerAct2 bimanual benchmark, MV-Actor achieves a state-of-the-art average success rate of 87.8\%. In real-world evaluations with more frequent viewpoint changes and unstable consumer-grade depth, MV-Actor outperforms both RGB and RGB-D baselines, further demonstrating the benefit of sharing semantic perception and reliable spatial awareness for bimanual manipulation.

\medskip
\noindent\textbf{Code:} \href{https://github.com/TianYinchen56/MV-Actor}{TianYinchen56/MV-Actor}
\begin{figure}[t]
    \centering
    \includegraphics[width=\columnwidth]{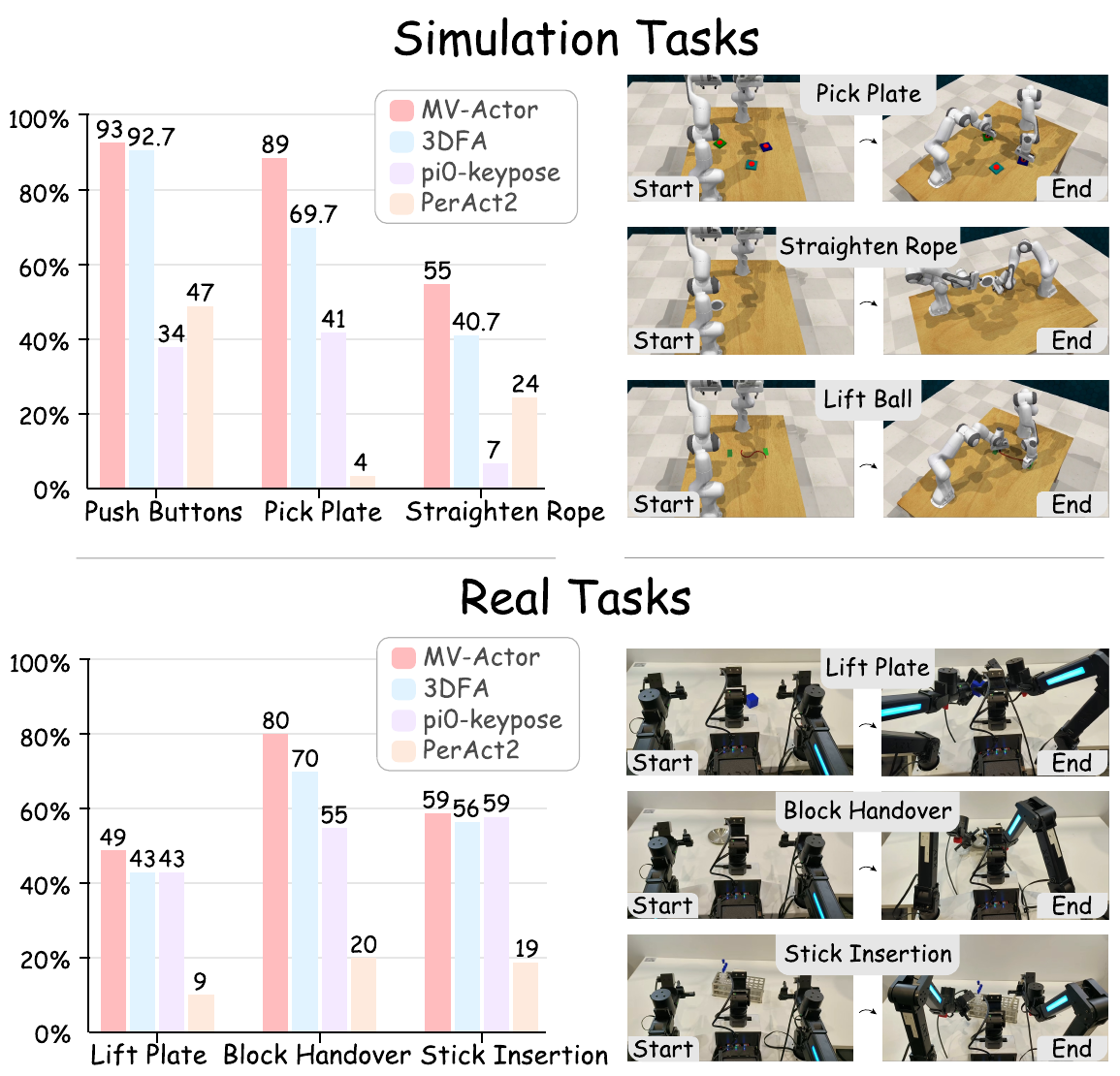}
    \caption{\textbf{Representative simulation and real-robot tasks.} Left: success rates (\%) on representative simulation and real-robot bimanual tasks. Right: start/end examples for the same tasks. MV-Actor achieves consistently best performance compared to existing methods.}
    \label{fig:teaser}
\end{figure}

\section{INTRODUCTION}
Robot manipulation plays a central role in scenarios such as industrial assembly~\cite{billard2019trends}. Existing single-arm methods typically rely on a single camera, so the policy perceives the scene from only one viewpoint~\cite{diffusionpolicy,rt1}, leaving the available visual information inherently limited. In contrast, bimanual systems are commonly equipped with multiple cameras, such as wrist-mounted and external views~\cite{act,fu2024mobile}. However, most bimanual policies often use these camera streams as separate visual inputs, so the perception is insufficiently shared across views.

To leverage these multi-view observations, existing methods construct multi-view representations in two main ways. The first category~\cite{act,rdt1b,pi0,bfa} encodes each view independently and fuses the resulting features at the policy level. This allows policies to receive observations from multiple cameras, but the correspondence between different views is not fully considered. The second category~\cite{peract,act3d,3ddiffuser,3dfa,li2026manivid} goes one step further by using sensor depth to lift visual features into a shared 3D space, such as voxels or point clouds. This shared spatial frame enables explicit multi-view alignment, but its reliability depends on depth sensor quality. Therefore, existing approaches either fuse views without multi-view association or rely on fragile depth-based alignment, leaving \textbf{sharing semantic perception} and \textbf{reliable spatial awareness} from multi-camera observations unresolved.

Based on the above insight, we propose MV-Actor, a multi-view perception framework for bimanual manipulation that constructs a unified semantic-spatial representation. On the semantic side, semantic perception is shared between features corresponding to the same physical region across different views, enabling visual tokens to draw semantic evidence from other cameras. On the spatial side, the rapid development of feed-forward reconstruction models has made it possible to obtain implicit spatial geometry priors from multi-view RGB images alone~\cite{dust3r,memix,pi3}. Our visual semantic features interact with spatial features produced by feed-forward reconstruction models, endowing the representation with reliable spatial awareness. Then the Head-Aware Routing Gate adaptively fuses the semantic and spatial refinements, with separate weights for the translation and rotation branches. Furthermore, to address the degradation of consumer-grade depth sensors, MV-Actor repairs depth information to provide reliable geometric support for Multi-view Semantic Interaction and reliable spatial awareness. In simulation experiments on the PerAct2 bimanual benchmark, MV-Actor achieves a state-of-the-art average success rate of 87.8\%. On a real bimanual platform, even under degraded consumer-grade depth conditions, MV-Actor outperforms voxel, point-cloud, and RGB-only baselines across three categories of bimanual tasks, with representative task outcomes shown in Fig.~\ref{fig:teaser}.
Our contributions can be summarized as follows:
\begin{itemize}
    \item We propose \textbf{MV-Actor}, a multi-view perception framework for bimanual manipulation. MV-Actor treats multi-camera observations as related views of the same scene and builds a unified representation that improves sharing semantic perception and reliable spatial awareness.
    \item We propose Multi-view Semantic Interaction to share semantic perception across physically matched regions, and Semantic-Spatial Token Interaction to enrich visual features with reliable spatial awareness from reconstruction priors. A depth repair module further recovers reliable metric depth from degraded depth sensor.
    \item We evaluate MV-Actor in simulation and on a real bimanual robot equipped with consumer-grade depth cameras. MV-Actor achieves state-of-the-art performance on PerAct2 (87.8\%) and outperforms both explicit 3D and RGB-only baselines in real-world experiments.
\end{itemize}
\begin{figure*}[t]
    \centering
    \includegraphics[width=\textwidth]{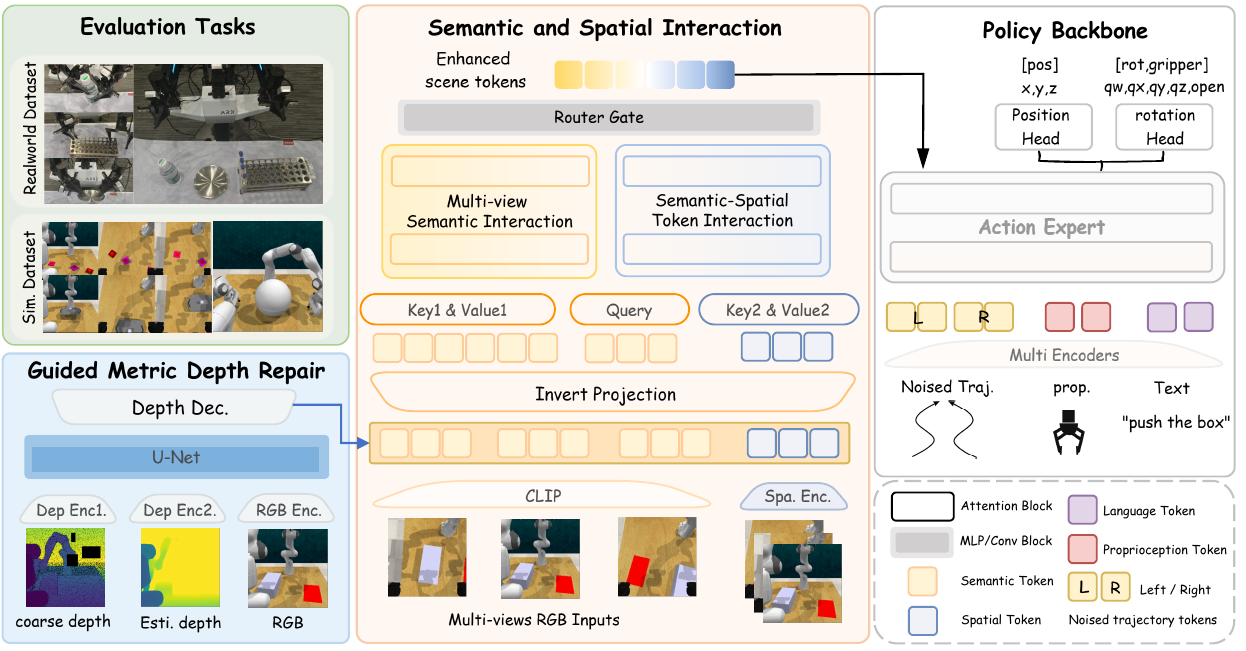}
    \caption{\textbf{MV-Actor overview.} Given multi-view RGB-D observations, MV-Actor first repairs degraded sensor depth by combining RGB texture and a Pi3X feed-forward depth prior, producing metric depth that can be back-projected into the robot frame. CLIP features provide semantic scene tokens, while Pi3X~\cite{pi3} provides feed-forward spatial tokens. Multi-view Semantic Interaction exchanges semantic evidence across physically corresponding regions, and Semantic-Spatial Token Interaction enables semantic tokens to acquire spatial awareness from the feed-forward spatial tokens. The Head-Aware Routing Gate forms enhanced scene tokens and routes them to position and rotation/gripper prediction streams before action decoding.}
    \label{fig:architecture}
\end{figure*}

\section{Related Work}
\subsection{Multi-View Perception}
Multi-view perception provides rich and complementary observations for robot scene understanding~\cite{mvmwm,reviwo,3dmvp,rvt,rvt2,bfa,peafowl}. Existing methods can be broadly divided into two categories.

The first category encodes each view independently or simply fuses feature from every single view. ACT~\cite{act}, $\pi_0$~\cite{pi0}, and RDT-1B~\cite{rdt1b} concatenate per-view tokens and feed them directly into the policy network. MVMwM~\cite{mvmwm}, ReViWo~\cite{reviwo}, and 3D-MVP~\cite{3dmvp} attend over per-view tokens to aggregate information. RVT~\cite{rvt} and RVT-2~\cite{rvt2} render virtual views to fuse multi-view features, BFA~\cite{bfa} selects best-view features, and PEAfowl~\cite{peafowl} aggregates local 3D neighbors by spatial proximity. A common limitation of these methods is that observations of the same physical region from different cameras are never explicitly associated, leaving sharing semantic perception across views underexploited.

The second category uses sensor depth to lift features into a shared 3D space. PerAct~\cite{peract} voxelizes the workspace and fuses multi-view features in a volumetric grid. Act3D~\cite{act3d}, 3D Diffuser Actor~\cite{3ddiffuser}, 3DFA~\cite{3dfa}, and ManiVID-3D~\cite{li2026manivid} lift features into point clouds or 3D representations for action prediction. These methods achieve multi-view fusion through spatial alignment, but depend heavily on depth sensor quality---consumer-grade cameras frequently produce holes and noise on transparent or reflective surfaces, degrading the resulting 3D representations. Recently, feed-forward reconstruction models such as DUSt3R~\cite{dust3r}, MeMix~\cite{memix}, and Pi3~\cite{pi3} can produce dense pointmaps and spatial tokens from multi-view RGB images alone, and related work in VLM and VLA~\cite{g2vlm,dynam3d,evo0} has shown that such spatial features improve spatial awareness in language-conditioned tasks. However, these priors have not yet been fully exploited in robot manipulation policies.

\subsection{Policy Learning for Bimanual Manipulation}
Bimanual manipulation requires coordinated motion and temporally precise interactions between two end-effectors, posing challenges beyond single-arm control. Imitation-learning methods remain the dominant paradigm: PerAct2~\cite{peract2} extends voxel-based policies to dual arms, VoxAct-B~\cite{voxactb} models dual-arm dependencies, InterACT~\cite{interact} introduces action chunking for bimanual coordination, RDT-1B~\cite{rdt1b} scales diffusion policies, and 3DFA~\cite{3dfa} generates actions via 3D flow matching~\cite{ddpm,ddim,dit}. In parallel, VLA models such as RT-1~\cite{rt1}, RT-2~\cite{rt2}, Octo~\cite{octo}, OpenVLA~\cite{openvla}, $\pi_0$~\cite{pi0}, and $\pi_{0.5}$~\cite{pi05} transfer pretrained visual-language knowledge into robot control, with generative action heads based on diffusion, flow matching, or autoregressive sequence modeling further improving action modeling~\cite{diffusionpolicy,flowmatching,zhang2025autoregressive}. While these architectures have advanced bimanual manipulation, most treat the multi-view perception front-end as a fixed input representation without exploring how to build richer perceptual features from multi-camera observations.

\subsection{Discussion}
In summary, existing multi-view manipulation policies mainly follow two paradigms. Independent-encoding methods concatenate or fuse per-view features inside the policy network, but they do not explicitly associate observations across views. 3D-lifting methods align views in an explicit geometric space, but their reliability is constrained by sensor-depth quality. PEAfowl~\cite{peafowl} is the closest prior, it also enhances multi-view VLA policies with geometry-aware cross-view aggregation. However, it builds cross-view representations through learned depth distributions and local lifted-neighbor aggregation, whereas MV-Actor explicitly shares semantic perception at calibrated physically corresponding regions and uses feed-forward reconstruction features to build reliable spatial awareness, and repairing degraded sensor depth to provide reliable geometric support.
\section{Method}
\subsection{Problem Formulation}
Let $\mathcal{V}=\{1,\dots,V\}$ denote the camera set. At time $t$, the policy receives multi-view RGB-D observations, a language instruction, and robot proprioception:
\begin{equation}
O_t = \left(\{I_t^{(v)}, D_t^{(v)}\}_{v\in \mathcal{V}}, \ell, q_t\right),
\end{equation}
where $I_t^{(v)}$ and $D_t^{(v)}$ denote the RGB image and depth map of view $v$, $\ell$ denotes the language instruction, and $q_t$ denotes proprioceptive state. For each view, frozen CLIP~\cite{clip} features are extracted as a $32\times32$ semantic token grid. Each token is back-projected into the robot frame:
\begin{equation}
p_i^{(v)} = T_{\mathrm{base} \leftarrow \mathrm{cam}}^{(v)} \cdot D^{(v)}(u_i) \cdot K^{(v)-1} \tilde{u}_i,
\end{equation}
where $K^{(v)}$ denotes camera intrinsics, $T_{\mathrm{base} \leftarrow \mathrm{cam}}^{(v)}$ is the camera-to-base extrinsic transform, and $\tilde{u}_i$ is the homogeneous pixel coordinate, forming a metric point cloud as the shared spatial carrier. In parallel, Pi3X~\cite{pi3} provides spatial tokens encoding multi-view geometric structure, and the language instruction is encoded by a frozen CLIP text encoder (Fig.~\ref{fig:architecture}). Based on these scene tokens, MV-Actor predicts an action chunk $A_t=\{a_\tau\}_{\tau=t}^{t+H-1}$, where each bimanual action is
\begin{equation}
a_\tau =
\left\{
(p_\tau^{\mathrm{left}}, r_\tau^{\mathrm{left}}, g_\tau^{\mathrm{left}}),
(p_\tau^{\mathrm{right}}, r_\tau^{\mathrm{right}}, g_\tau^{\mathrm{right}})
\right\}.
\end{equation}
Here $p$, $r$, and $g$ denote end-effector position, rotation, and gripper state.

\subsection{Multi-view and Semantic-Spatial Token Interaction}
Environmental variations such as occlusion and viewpoint changes can cause multi-view observations to produce inconsistent semantic responses across cameras, and 2D semantic tokens alone lack spatial awareness. MV-Actor therefore introduces two parallel interaction paths in the metric point-cloud space: \textbf{Multi-view Semantic Interaction} propagates semantic evidence across physically corresponding regions, while \textbf{Semantic-Spatial Token Interaction} enables semantic tokens to query feed-forward spatial tokens. These two paths produce semantic and spatial residuals, which are later routed to downstream action heads.

\begin{figure}[t]
    \centering
    \includegraphics[width=\linewidth]{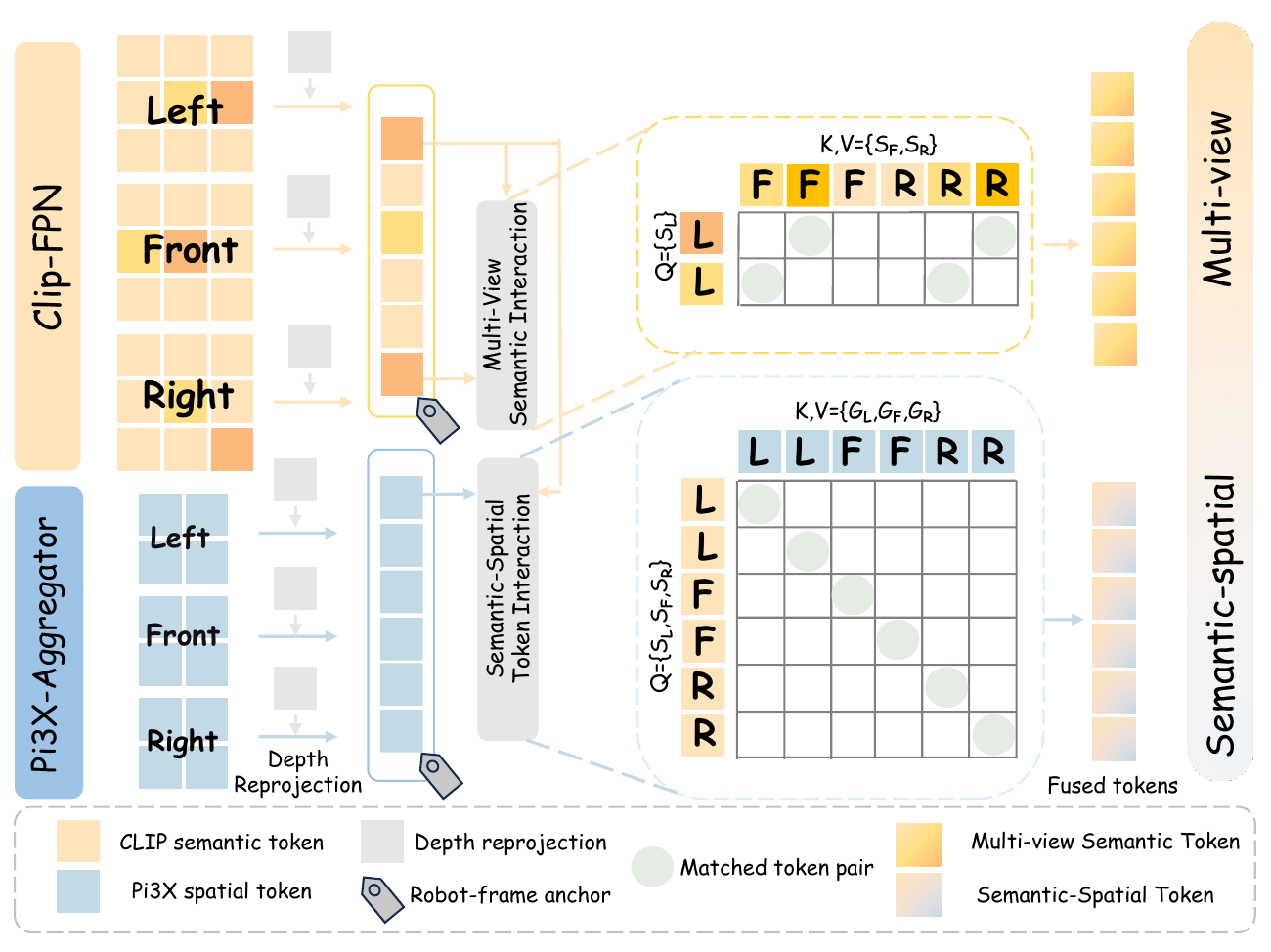}
    \caption{\textbf{Multi-view and semantic-spatial token interaction.} MV-Actor builds semantic tokens from CLIP-FPN and spatial tokens from the Pi3X aggregator. The upper branch performs Multi-view Semantic Interaction: semantic queries retrieve key/value tokens from reprojection-consistent regions in other calibrated views. The lower branch performs Semantic-Spatial Token Interaction: semantic queries attend to Pi3X spatial tokens to inject spatial awareness. The two residuals are fused into the final scene tokens.}
    \label{fig:sem2sem_method}
\end{figure}

\paragraph{Multi-view Semantic Interaction}
When a target object is partially hidden by an arm or another object, the semantic token from a single view may miss crucial evidence. Multi-view Semantic Interaction uses calibrated reprojection to search for evidence from other views. As shown by the upper branch in Fig.~\ref{fig:sem2sem_method}, it avoids direct 3D KNN aggregation over the entire point cloud, which can introduce neighbors from nearby but physically different surfaces. Instead, it establishes reprojection-consistent correspondences so that semantic propagation happens only between regions that refer to the same physical location.

\textit{Locating multi-view correspondences.}
For a semantic token in view $v$ with 3D anchor $p_i^{(v)}$, we project the anchor into a target view $u$ on the semantic-token grid:
\begin{equation}
\hat{x}_u=R_u p_i^{(v)}+t_u,\quad
\hat{z}_u=[\hat{x}_u]_z,\quad
\hat{c}_u=\Pi\!\left(K_u^{\mathrm{grid}}\hat{x}_u\right),
\end{equation}
where $(R_u,t_u)$ transforms robot-frame points into the camera frame of view $u$, $K_u^{\mathrm{grid}}$ is the camera intrinsic matrix scaled from image resolution to the $32\times32$ token grid, $\Pi([x,y,z]^\top)=(x/z,y/z)$ denotes perspective normalization, and $\hat{z}_u$ is the camera-frame depth.

\textit{Spatially consistent candidate filtering.}
Candidate tokens are searched in a radius-one local window around $\hat{c}_u$ on the $32\times32$ semantic-token grid. A candidate token $j$ in view $u$ is valid only if it satisfies both grid-plane alignment and depth consistency:
\begin{align}
  \| \hat{c}_u - c_j^{(u)} \|_2 &\le \epsilon_{\mathrm{reproj}}, \label{eq:reproj_en} \\
  | \hat{z}_u - z_j^{(u)} | &\le \epsilon_{\mathrm{depth}}, \label{eq:depth_en}
\end{align}
where $c_j^{(u)}$ is the candidate center in token-grid coordinates, $\epsilon_{\mathrm{reproj}}$ is measured in token-grid cells, and $z_j^{(u)}$ is the $z$ coordinate of its repaired-depth 3D anchor after transformation into the camera frame of view $u$. Eq.~\eqref{eq:reproj_en} enforces local grid alignment, while Eq.~\eqref{eq:depth_en} rejects perspective-overlap artifacts where two different 3D surfaces project to a similar image location.

Among the valid candidates in each target view, we retain the token with the smallest combined reprojection-and-depth error (Eqs.~\eqref{eq:reproj_en}--\eqref{eq:depth_en}); the retained matches form the correspondence set $\mathcal{C}_i^{(v)}$. If no candidate satisfies the constraints in a target view, a validity mask removes that view from subsequent attention. This strict physical filtering is the key to stable Multi-view Semantic Interaction.

\textit{Local attention aggregation.}
Given $\mathcal{C}_i^{(v)}$, Multi-view Semantic Interaction updates the current token with a masked local attention residual. Let $d_h=d/N_h$ be the per-head dimension. We write the multi-head attention in compact form and omit the head index for readability:
\begin{equation}
\alpha_{i,uj}
=
\operatorname{softmax}_{\mathcal{C}_i^{(v)}}\!
\left(
\frac{(W_Q s_i^{(v)})^\top W_K s_j^{(u)}}{\sqrt{d_h}}
\right),
\end{equation}
\begin{equation}
\Delta s_i^{\mathrm{sem}}
=
\lambda_s W_O
\sum_{(u,j)\in\mathcal{C}_i^{(v)}}
\alpha_{i,uj} W_V s_j^{(u)}.
\end{equation}
The softmax is normalized only over valid candidates; a token with no valid match receives a zero residual. We apply this block twice with $N_h=8$ heads and $\lambda_s=0.1$.

\paragraph{Semantic-Spatial Token Interaction}
2D semantic tokens alone cannot represent local 3D structure. As shown by the lower branch in Fig.~\ref{fig:sem2sem_method}, Semantic-Spatial Token Interaction enables each semantic token to attend to an aligned feed-forward spatial token for spatial awareness.

Pi3X~\cite{pi3} spatial tokens from each view are projected to dimension $d$, layer-normalized, and resized to the $32\times32$ semantic-token grid via nearest-neighbor copying. For view $v$, let $U^{(v)}$ denote its feed-forward spatial-token grid. The aligned spatial token for semantic token $s_i^{(v)}$ at location $c_i^{(v)}$ is $\tilde{g}_i^{(v)}=\operatorname{Align}(\operatorname{LN}(W_S U^{(v)}),c_i^{(v)})$. Each semantic token then attends to its aligned spatial token through a per-head sigmoid-gated residual:
\begin{equation}
\beta_{i,h}^{(v)} =
\sigma\!
\left(
\frac{
(W_{Q,h}^{\mathrm{spa}} s_i^{(v)})^\top
W_{K,h}^{\mathrm{spa}}\tilde{g}_i^{(v)}
}{\sqrt{d_h}}
\right),
\end{equation}
\begin{equation}
\Delta s_i^{\mathrm{spa},(v)}
=
\lambda_{\mathrm{spa}} W_O^{\mathrm{spa}}
\operatorname{Concat}_{h}
\left(\beta_{i,h}^{(v)} W_{V,h}^{\mathrm{spa}}\tilde{g}_i^{(v)}\right),
\end{equation}
where $\beta_{i,h}^{(v)}$ is a scalar gate for head $h$, broadcast over the head dimension, and $\lambda_{\mathrm{spa}}=0.1$.

\paragraph{Parallel residual fusion}
Multi-view Semantic Interaction and Semantic-Spatial Token Interaction act in parallel on the shared base scene tokens $s=\{s_i\}$. They produce complementary residuals: $\Delta s^{\mathrm{sem}}$ from multi-view semantic propagation and $\Delta s^{\mathrm{spa}}$ from local spatial querying. Keeping these residuals separate is important because position and rotation heads may prefer different mixtures of semantic and spatial evidence. The next subsection describes how this mixture is controlled.

\subsection{Head-Aware Routing Gate}
Position and rotation predictions depend on the semantic and spatial residuals to different degrees. We introduce a lightweight \textbf{Head-Aware Routing Gate} that adaptively controls the mixture.

The Head-Aware Routing Gate takes as input the base scene tokens $s$, instruction tokens $e_\ell$, proprioception $q_t$, and the two residuals $\Delta s^{\mathrm{sem}}$, $\Delta s^{\mathrm{spa}}$. It forms a context vector:
\begin{equation}
z_{\mathrm{route}} =
\begin{aligned}[t]
\Big[
&\operatorname{Pool}(s), \operatorname{Pool}(e_\ell), \psi(q_t), \\
&\operatorname{Pool}(\Delta s^{\mathrm{sem}}),
\operatorname{Pool}(\Delta s^{\mathrm{spa}})
\Big],
\end{aligned}
\end{equation}
where $\operatorname{Pool}(\cdot)$ denotes mean pooling and $\psi$ projects proprioception to the latent dimension. A two-layer MLP predicts four sigmoid gates:
\begin{equation}
\gamma = \sigma(\operatorname{MLP}_{\mathrm{route}}(z_{\mathrm{route}})),
\end{equation}
\begin{equation}
s^h = s + \gamma_{\mathrm{sem}}^h \Delta s^{\mathrm{sem}}
+ \gamma_{\mathrm{spa}}^h \Delta s^{\mathrm{spa}},
\quad h\in\{\mathrm{pos},\mathrm{rot}\}.
\end{equation}
Here $\gamma=(\gamma_{\mathrm{sem}}^{\mathrm{pos}},\gamma_{\mathrm{spa}}^{\mathrm{pos}},\gamma_{\mathrm{sem}}^{\mathrm{rot}},\gamma_{\mathrm{spa}}^{\mathrm{rot}})$ are scalar gates broadcast over tokens and channels. The gripper head shares the rotation-routed representation.

\subsection{Guided Metric Depth Repair}
The policy backbone encodes spatial structure via 3D Rotary Position Encoding (RoPE), which requires a metric 3D coordinate for every visual token. We obtain this coordinate by back-projecting through the repaired depth:
\begin{equation}
\hat{p}_i^{(v)}
=
T_{\mathrm{base}\leftarrow\mathrm{cam}}^{(v)}
d_{\mathrm{out}}^{(v)}(u_i)
K^{(v)-1}\tilde{u}_i .
\label{eq:repaired_backproj}
\end{equation}
Consumer-grade depth sensors produce holes and noisy boundaries on reflective or transparent surfaces, yielding incorrect 3D positions that distort the positional encoding and corrupt spatial reasoning. We therefore repair the sensor depth before back-projection.

The repair module fuses three complementary inputs: (i) the raw sensor depth $d_{\mathrm{noisy}}^{(v)}$, which preserves metric scale; (ii) the RGB image $I^{(v)}$, which provides sharp boundary cues; and (iii) a Pi3X feed-forward depth prior, which fills missing regions with cross-view consistent geometry:
\begin{equation}
\left(P_{\pi^3}^{(v)}, d_{\pi^3}^{(v)}\right)
=
\mathcal{G}_{\pi^3}\!\left(\{I^{(u)}\}_{u\in\mathcal{V}}\right)^{(v)}.
\label{eq:pi3_depth}
\end{equation}
Since $d_{\pi^3}^{(v)}$ may exhibit scale drift, it serves as geometric guidance rather than the final output.

We adopt a three-branch U-Net~\cite{unet} encoder--decoder. At scale $l$, each branch encodes one input, and a fusion block merges them with the coarser-scale decoder feature:
\begin{equation}
F_l =
\Phi_l\!\left(
E_l^{d}(d_{\mathrm{noisy}}^{(v)}),
E_l^{I}(I^{(v)}),
E_l^{\pi}(d_{\pi^3}^{(v)}),
F_{l+1}
\right).
\label{eq:depth_fusion}
\end{equation}
The decoder produces a non-negative repaired depth:
\begin{equation}
d_{\mathrm{out}}^{(v)} = \max\!\left(0, H_{\mathrm{depth}}(F_0)\right).
\label{eq:depth_output}
\end{equation}
The resulting depth retains the sensor's metric scale, recovers missing geometry from the feed-forward depth prior, and preserves object boundaries via RGB guidance. It is then used in Eq.~\eqref{eq:repaired_backproj} to build the robot-frame point cloud for all subsequent token interactions and action prediction.

\subsection{Policy Backbone and Training Objective}
MV-Actor uses a 3D denoising policy backbone conditioned on scene tokens, instruction tokens, proprioception, and noisy action trajectories. The action decoder predicts bimanual end-effector pose updates and gripper states from noisy action inputs. The scene encoder described above supplies routed scene tokens for position and rotation prediction streams.

Positions are represented as workspace-normalized 3D coordinates, rotations are converted from quaternions to a continuous 6D representation~\cite{rot6d}, and gripper states are supervised as binary open/close labels. For the denoising targets, we use element-wise mean L1 losses for position and rotation and binary cross-entropy for the gripper state:
\begin{align}
\mathcal{L}_{\mathrm{pos}} &= \operatorname{mean}\left|\hat{\epsilon}_{\mathrm{pos}}-\epsilon^\star_{\mathrm{pos}}\right|,\\
\mathcal{L}_{\mathrm{rot}} &= \operatorname{mean}\left|\hat{\epsilon}_{\mathrm{rot}}-\epsilon^\star_{\mathrm{rot}}\right|,\\
\mathcal{L}_{\mathrm{grip}} &= \operatorname{BCEWithLogits}\!\left(\hat{g}, g^\star\right).
\end{align}
Here $\epsilon$ denotes the flow/denoising target rather than the action itself. The total objective is
\begin{equation}
\mathcal{L}
= \lambda_{\mathrm{pos}}\mathcal{L}_{\mathrm{pos}}
+ \lambda_{\mathrm{rot}}\mathcal{L}_{\mathrm{rot}}
+ \mathcal{L}_{\mathrm{grip}}.
\end{equation}
We set the loss weights to $\lambda_{\mathrm{pos}}=30$ and $\lambda_{\mathrm{rot}}=10$.
\section{Experiments}
We evaluate MV-Actor in simulation on the PerAct2 bimanual benchmark. PerAct2 contains 13 bimanual manipulation tasks requiring coordination, contact reasoning, and spatial precision. We use three calibrated RGB-D views: front, left wrist, and right wrist. Unless otherwise noted, images and depth maps are resized to $256\times256$.

\subsection{Simulation Experiments}
\subsubsection{Simulation Environment Setup}

\paragraph{Metrics}
For simulation, we report online task success rate and offline action prediction metrics, including position error, position accuracy, rotation error, rotation accuracy, and gripper accuracy.

\paragraph{Implementation details}
Ablation experiments are trained for 200k updates, while the full MV-Actor is trained for 400k updates, taking about 20 hours on eight A100 GPUs.

\paragraph{Baselines}
We compare against explicit 3D RGB-D policies, RGB-only policies, and our repaired-depth RGB-D setting.

\subsubsection{Simulation Results}

\begin{table*}[!t]
    \centering
    \caption{\textbf{Main PerAct2 comparison.} Except for Ours, results are organized from the 3DFA Table~1~\cite{3dfa}. Bold and underline denote the best and second-best result in each task column. Gray background marks MV-Actor. Missing entries indicate unreported results. }
    \label{tab:main_results}
    \scriptsize
    \setlength{\tabcolsep}{0pt}
    \renewcommand{\arraystretch}{1.08}
    \begin{tabular}{@{}p{0.17\textwidth}p{0.20\textwidth}*{7}{>{\centering\arraybackslash}p{0.085\textwidth}}@{}}
        \toprule
        Method & Category & Avg. & PB & LB & DB & PP & ID & BF \\
        \midrule
        RVT-LF~\cite{rvt}        & \multirow{8}{*}{\mbox{RGB-D explicit 3D}} & 10.5 & 52.0 & 17.0 & 39.0 & 3.0 & 10.0 & 0.0 \\
        PerAct-LF~\cite{peract}     & & 17.5 & 57.0 & 40.0 & 10.0 & 2.0 & 27.0 & 0.0 \\
        PerAct2~\cite{peract2}       & & 16.8 & 6.0 & 50.0 & 47.0 & 4.0 & 10.0 & 3.0 \\
        DP3~\cite{dp3}           & & 25.6 & 56.0 & 64.0 & \pending & \pending & \pending & \pending \\
        KStarDiffuser~\cite{kstar} & & 68.3 & 83.0 & 98.7 & \pending & \pending & \pending & \pending \\
        PPI~\cite{ppi}           & & 80.8 & \best{96.7} & 89.3 & \pending & \pending & 79.7 & \pending\\
        AnyBimanual~\cite{anybimanual}   & & 32.0 & 46.0 & 36.0 & 73.0 & 8.0 & \pending & 26.0 \\
        3DFA~\cite{3dfa}          & & \second{85.1} & 92.7 & \second{99.7} & \second{92.7} & \second{69.7} & \second{93.0} & \second{89.3} \\
        \midrule
        ACT~\cite{act}             & \multirow{2}{*}{\mbox{RGB-only}} & 5.9 & 0.0 & 36.0 & 4.0 & 0.0 & 13.0 & 0.0 \\
        $\pi_0$-keypose~\cite{pi0} & & 43.7 & \second{93.0} & 97.0 & 38.0 & 41.0 & 40.0 & 22.0 \\
        \midrule
        \oursbar{\textbf{Ours}}{RGB-D repaired depth}{\best{87.8}}{89.0}{\best{100.0}}{\best{93.0}}{\best{89.0}}{\best{95.0}}{\best{90.0}} \\
        \bottomrule
    \end{tabular}

    \vspace{4pt}

    \begin{tabular}{@{}p{0.17\textwidth}p{0.20\textwidth}*{7}{>{\centering\arraybackslash}p{0.085\textwidth}}@{}}
        \toprule
        Method & Category & HO & PL & SR & SW & LT & HE & TO \\
        \midrule
        RVT-LF~\cite{rvt}        & \multirow{8}{*}{\mbox{RGB-D explicit 3D}} & 0.0 & 3.0 & 3.0 & 0.0 & 6.0 & 0.0 & 3.0 \\
        PerAct-LF~\cite{peract}     & & 0.0 & 11.0 & 21.0 & 28.0 & 14.0 & 9.0 & 8.0 \\
        PerAct2~\cite{peract2}       & & 11.0 & 12.0 & 24.0 & 0.0 & 1.0 & 41.0 & 9.0 \\
        DP3~\cite{dp3}           & & \pending & 6.3 & \pending & 1.7 & \pending & 0.0 & \pending \\
        KStarDiffuser~\cite{kstar} & & \pending & 43.7 & \pending & 89.0 & \pending & 27.0 & \pending \\
        PPI~\cite{ppi}           & & \pending & 46.3 & \pending & \second{98.7} & 92.0 & 62.7 & \pending \\
        AnyBimanual~\cite{anybimanual}   & & 15.0 & 7.0 & 24.0 & 67.0 & 14.0 & 44.0 & 24.0 \\
        3DFA~\cite{3dfa}          & & \second{89.0} & \best{74.0} & \second{40.7} & \best{99.3} & \best{94.7} & \best{96.0} & \best{94.7} \\
        \midrule
        ACT~\cite{act}             & \multirow{2}{*}{\mbox{RGB-only}} & 0.0 & 0.0 & 16.0 & 0.0 & 6.0 & 0.0 & 2.0 \\
        $\pi_0$-keypose~\cite{pi0} & & 2.0 & 27.0 & 7.0 & 2.0 & 72.0 & 59.0 & 68.0 \\
        \midrule
        \oursbar{\textbf{Ours}}{RGB-D repaired depth}{\best{92.0}}{\second{64.0}}{\best{55.0}}{95.0}{\second{93.0}}{\second{94.0}}{\second{93.0}} \\
        \bottomrule
    \end{tabular}
\end{table*}

Table~\ref{tab:main_results} compares MV-Actor with prior methods on PerAct2. We group methods by representation form: RGB-only implicit multi-view fusion, explicit RGB-D 3D policies, and our metric-space semantic-spatial scene-token representation. Baseline results are cited from 3DFA~\cite{3dfa}. Bold and underline denote best and second-best.

MV-Actor achieves an average success rate of 87.8\%, substantially outperforming the RGB-only baselines ACT and $\pi_0$-keypose and exceeding the strongest RGB-D method, 3DFA (85.1\%).

Unlike other methods that process each view independently, MV-Actor explicitly aggregates complementary semantic evidence across views at physically corresponding regions. On tasks such as \textit{lift\_ball} and \textit{push\_buttons}, where three calibrated views share substantial spatial overlap, more physically corresponding regions are available for Multi-view Semantic Interaction; the aggregated multi-angle semantic cues form a richer and more consistent target representation, leading to near-perfect success rates.

On tasks demanding high spatial precision, such as \textit{pick\_plate} (+19.3\% over 3DFA) and \textit{straighten\_rope} (+14.3\%), Semantic-Spatial Token Interaction injects geometric awareness from feed-forward reconstruction features into semantic tokens; the enriched tokens encode not only what the target object is but also where it is in metric 3D space, enabling the policy to position end-effectors with higher accuracy. This observation is consistent with recent evidence that view-invariant robot representations improve spatial reasoning in manipulation~\cite{li2026manivid}. In \textit{pick\_plate}, spatial awareness helps locate feasible grasp regions across different edges captured by different cameras; in \textit{straighten\_rope}, it enables accurate endpoint reasoning along the full rope configuration observed from multiple angles.

On long-horizon tasks such as \textit{sweep\_dust}, \textit{lift\_tray}, \textit{handover\_easy}, \textit{take\_tray\_out\_of\_oven}, and \textit{pick\_up\_laptop}, MV-Actor slightly underperforms 3DFA because extended manipulation introduces large viewpoint variation that reduces multi-view overlap and degrades feed-forward reconstruction quality, weakening both Multi-view Semantic Interaction and spatial awareness.

\begin{figure}[!t]
    \centering
    \includegraphics[width=\linewidth]{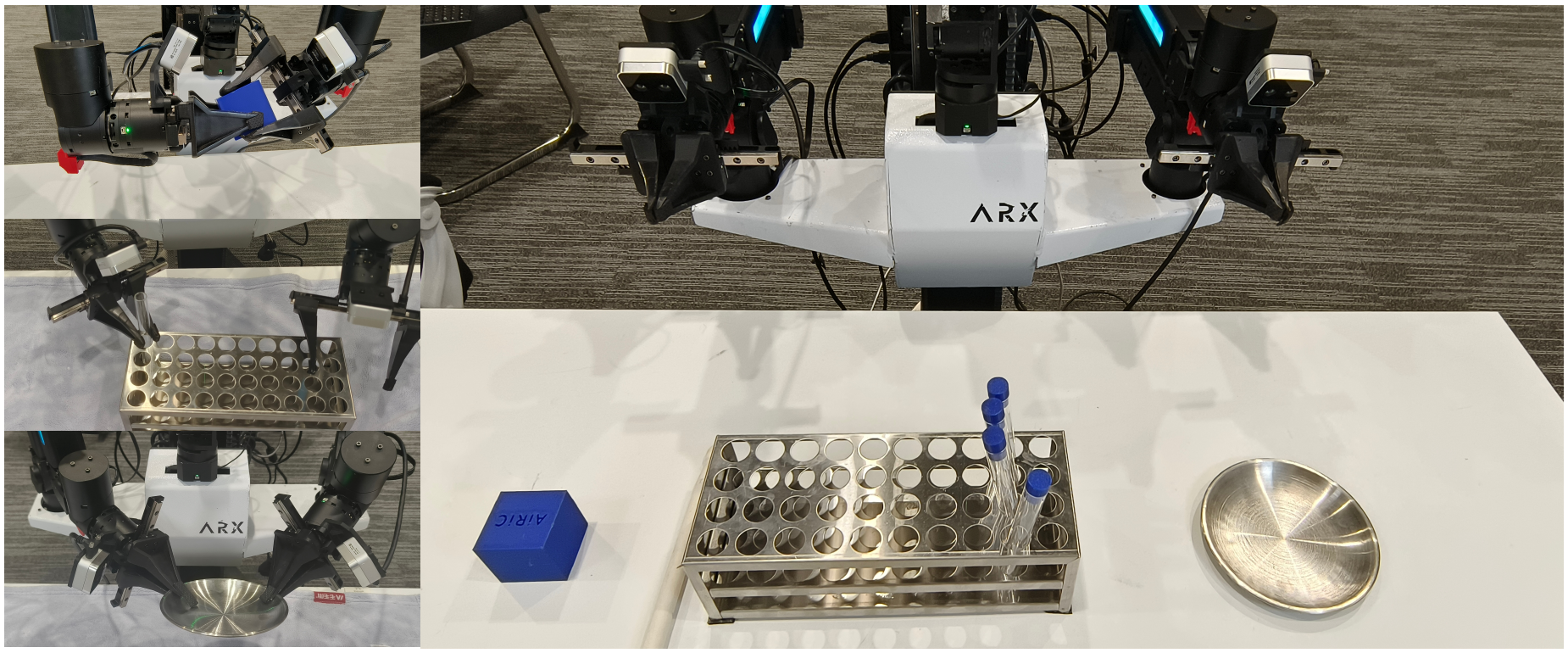}
    \caption{\textbf{ARK-Lift real-robot setup and tasks.} The platform uses three calibrated RGB-D cameras. The task examples cover cooperative lifting, stick insertion, and contact-based handover.}
    \label{fig:real_robot_setup}
\end{figure}

\subsection{Real-Robot Experiments}
\begin{table}[!t]
    \centering
    \caption{REAL-WORLD STAGED SCORES (\%) ON ARK-LIFT OVER 10 TRIALS PER TASK. BOLD DENOTES THE BEST RESULT.}
    \label{tab:real_world_results}
    \scriptsize
    \setlength{\tabcolsep}{0pt}
    \renewcommand{\arraystretch}{1.08}
    \begin{tabular}{@{}>{\raggedright\arraybackslash}p{0.30\columnwidth}*{4}{>{\centering\arraybackslash}p{0.175\columnwidth}}@{}}
        \toprule
        Method & Avg. & \makecell[c]{Lift\\Plate} & \makecell[c]{Block\\Handover} & \makecell[c]{Stick\\Insertion} \\
        \midrule
        PerAct2~\cite{peract2} & 16.6 & 9.9 & 20.0 & 19.8 \\
        3DFA~\cite{3dfa} & 58.1 & 43.0 & 75.0 & 56.2 \\
        $\pi_0$~\cite{pi0} & 52.5 & 43.0 & 55.0 & 59.4 \\
        \midrule
        \realoursbar{\textbf{MV-Actor (Ours)}}{\textbf{63.1}}{\textbf{49.7}}{\textbf{80.0}}{\textbf{59.6}} \\
        \bottomrule
    \end{tabular}
\end{table}
\subsubsection{Real-Robot Setup}
We further evaluate MV-Actor on the ARK-Lift bimanual robot platform shown in Fig.~\ref{fig:real_robot_setup} to test real-world deployment under depth degradation and arm occlusion.

\paragraph{Hardware platform}
The real setup uses three Intel RealSense D405 RGB-D cameras mounted as front, left wrist, and right wrist views. Input images are resized to $256\times256$. All cameras are hand-eye calibrated and aligned with the robot base frame. Policy inference runs on a single RTX 4090 GPU.

\paragraph{Data collection and scoring}
We evaluate three representative bimanual tasks: \textbf{Lift Plate}, \textbf{Block Handover}, and \textbf{Stick Insertion}. For each task, we collect 100 training demonstrations and evaluate 10 independent test trials. Two-stage tasks assign 0.50 to each stage; three-stage tasks use cumulative milestones of 0.33, 0.66, and 1.00. The scenes include reflective tables, self-occlusion by both arms, tool occlusion, and local depth holes.

\paragraph{Compared methods}
We compare four representative policies under the same task definitions and scoring protocol: PerAct2~\cite{peract2} as a voxel-based policy, 3DFA~\cite{3dfa} as a point-cloud policy, $\pi_0$~\cite{pi0} as an RGB-only VLA model, and MV-Actor as a repaired-depth semantic-spatial policy.

\subsubsection{Real-World Results}

Table~\ref{tab:real_world_results} compares four policies on the ARK-Lift platform across three bimanual tasks. Bold denotes the best result.

MV-Actor obtains the highest average staged score of 63.1\%, improving over 3DFA, $\pi_0$, and PerAct2 by 5.0, 10.6, and 46.5 percentage points, respectively. On Block Handover, MV-Actor reaches 80.0\%, outperforming 3DFA (75.0\%) and $\pi_0$ (55.0\%). In this task, the block is frequently occluded by grippers and moves between arms. Multi-view semantic consistency and the spatial awareness introduced by Semantic-Spatial Token Interaction are particularly effective in this setting.

On Lift Plate, MV-Actor improves over both 3DFA and $\pi_0$ by 6.7 points. This task uses a metallic plate, whose reflective surface introduces many noisy depth points and local holes in the D405 observations. Directly back-projecting such raw depth often yields an unstable point cloud around the plate boundary and feasible contact regions. The depth repair module combines RGB boundary cues with the Pi3X feed-forward depth prior to suppress depth noise and recover a more continuous metric spatial carrier, which helps estimate the two grippers' support positions relative to the plate. On Stick Insertion, 3DFA, $\pi_0$, and MV-Actor obtain similar scores because the task requires accurately grasping a thin stick and inserting it into a test-tube rack. The manipulation horizon is longer, multi-view overlap is limited, and success is highly sensitive to spatial precision.

\begin{table*}[!t]
    \centering
    \caption{\textbf{Interaction module ablation.} We incrementally add each component and report offline metrics at 50k, 100k, and 200k steps. Bold/underline mark best/second-best, and colored parentheses show change from baseline.}
    \label{tab:interaction_ablation_updated}
    \small
    \setlength{\tabcolsep}{2.4pt}
    \renewcommand{\arraystretch}{1.12}
    \begin{tabular*}{\textwidth}{@{\extracolsep{\fill}}cccccccccc@{}}
        \toprule
        Configuration & Multi-view & Sem-spa & \makecell{Head-Aware\\Gate} & Step & pos\_l2$\downarrow$ & pos\_acc~(\%)$\uparrow$ & rot\_l1$\downarrow$ & rot\_acc~(\%)$\uparrow$ & gripper~(\%)$\uparrow$ \\
        \midrule
        \multirow{3}{*}{Baseline} & \multirow{3}{*}{\xmark} & \multirow{3}{*}{\xmark} & \multirow{3}{*}{\xmark} & 50k & 0.021 & 35.8 & 0.072 & 35.3 & 98.0 \\
         & & & & 100k & 0.016 & 60.6 & 0.054 & 55.4 & \asecond{98.2} \\
         & & & & 200k & 0.012 & 81.0 & \asecond{0.046} & 70.2 & \abest{98.3} \\
        \midrule
        \multirow{3}{*}{+ Multi-view} & \multirow{3}{*}{\cmark} & \multirow{3}{*}{\xmark} & \multirow{3}{*}{\xmark} & 50k & \asecond{0.020}\gain{+0.001} & 37.0\gain{+1.2} & \asecond{0.069}\gain{+0.003} & 35.4\gain{+0.1} & \asecond{98.1}\gain{+0.1} \\
         & & & & 100k & 0.017\loss{-0.001} & 53.3\loss{-7.3} & \asecond{0.053}\gain{+0.001} & \asecond{60.2}\gain{+4.8} & 98.1\loss{-0.1} \\
         & & & & 200k & \asecond{0.012}\same{+0.000} & 74.1\loss{-6.9} & \abest{0.045}\gain{+0.001} & 69.9\loss{-0.3} & \asecond{98.2}\loss{-0.1} \\
        \midrule
        \multirow{3}{*}{+ Sem-spa} & \multirow{3}{*}{\xmark} & \multirow{3}{*}{\cmark} & \multirow{3}{*}{\xmark} & 50k & \asecond{0.020}\gain{+0.001} & \asecond{42.2}\gain{+6.4} & 0.073\loss{-0.001} & \asecond{39.0}\gain{+3.7} & 98.0\same{+0.0} \\
         & & & & 100k & \asecond{0.015}\gain{+0.001} & \asecond{70.0}\gain{+9.4} & 0.056\loss{-0.002} & 56.8\gain{+1.4} & 98.0\loss{-0.2} \\
         & & & & 200k & 0.012\same{+0.000} & \asecond{82.6}\gain{+1.6} & 0.050\loss{-0.004} & \abest{71.0}\gain{+0.8} & 97.9\loss{-0.4} \\
        \midrule
        \multirow{3}{*}{+ Both (no router)} & \multirow{3}{*}{\cmark} & \multirow{3}{*}{\cmark} & \multirow{3}{*}{\xmark} & 50k & 0.021\same{+0.000} & 33.1\loss{-2.7} & 0.067\gain{+0.005} & 43.3\gain{+8.0} & 98.0\same{+0.0} \\
         & & & & 100k & 0.016\same{+0.000} & 60.6\same{+0.0} & \asecond{0.053}\gain{+0.001} & 57.9\gain{+2.5} & \asecond{98.2}\same{+0.0} \\
         & & & & 200k & 0.013\loss{-0.001} & 79.3\loss{-1.7} & \asecond{0.046}\same{+0.000} & 70.8\gain{+0.6} & \asecond{98.2}\loss{-0.1} \\
        \midrule
        \multirow{3}{*}{MV-Actor (full)} & \multirow{3}{*}{\cmark} & \multirow{3}{*}{\cmark} & \multirow{3}{*}{\cmark} & 50k & \abest{0.015}\gain{+0.006} & \abest{56.8}\gain{+21.0} & \abest{0.063}\gain{+0.009} & \abest{40.2}\gain{+4.9} & \abest{98.2}\gain{+0.2} \\
         & & & & 100k & \abest{0.012}\gain{+0.004} & \abest{73.7}\gain{+13.1} & \abest{0.052}\gain{+0.002} & \abest{63.1}\gain{+7.7} & \abest{98.3}\gain{+0.1} \\
         & & & & 200k & \abest{0.009}\gain{+0.003} & \abest{86.3}\gain{+5.3} & \abest{0.045}\gain{+0.001} & \asecond{70.6}\gain{+0.4} & \abest{98.3}\same{+0.0} \\
        \bottomrule
    \end{tabular*}
\end{table*}

\subsection{Ablation Studies}
We conduct two ablation studies. The first examines how Multi-view Semantic Interaction, Semantic-Spatial Token Interaction, and the Head-Aware Routing Gate affect action prediction. The second analyzes how different depth guidance signals affect the repaired depth used for robot-frame 3D token anchoring.

\subsubsection{Cross-View and Semantic-Spatial Interaction}
Table~\ref{tab:interaction_ablation_updated} compares five interaction configurations---baseline (no interaction), adding Multi-view Semantic Interaction alone, adding Semantic-Spatial Token Interaction alone, combining both without the Head-Aware Routing Gate, and the full design with the Head-Aware Routing Gate---on offline action prediction metrics at 50k, 100k, and 200k training steps. Bold and underline denote best and second-best; colored parentheses show change relative to the baseline.

\begin{figure}[!t]
    \centering
    \includegraphics[width=\linewidth]{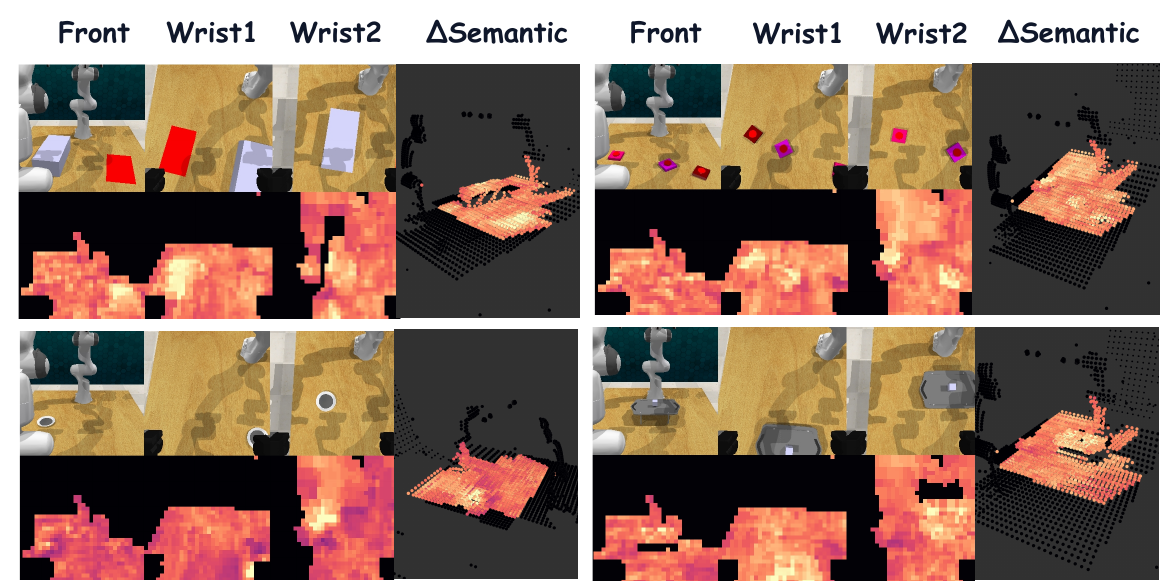}
    \caption{\textbf{Multi-view Semantic Interaction visualization.} Heatmaps show the magnitude of semantic feature updates after Multi-view Semantic Interaction. High responses concentrate on task-relevant regions, indicating that semantic evidence is exchanged around physically corresponding multi-view regions.}
    \label{fig:sem2sem_vis}
\end{figure}

Multi-view Semantic Interaction mainly improves rotation prediction. At 100k and 200k steps, $\mathrm{rot\_l1}$ drops to 0.053 and 0.045, the best at each step. Fig.~\ref{fig:sem2sem_vis} shows that feature updates concentrate on task-relevant regions, confirming that the gains come from stronger multi-view local semantic consistency.

\begin{figure}[!t]
    \centering
    \includegraphics[width=\linewidth]{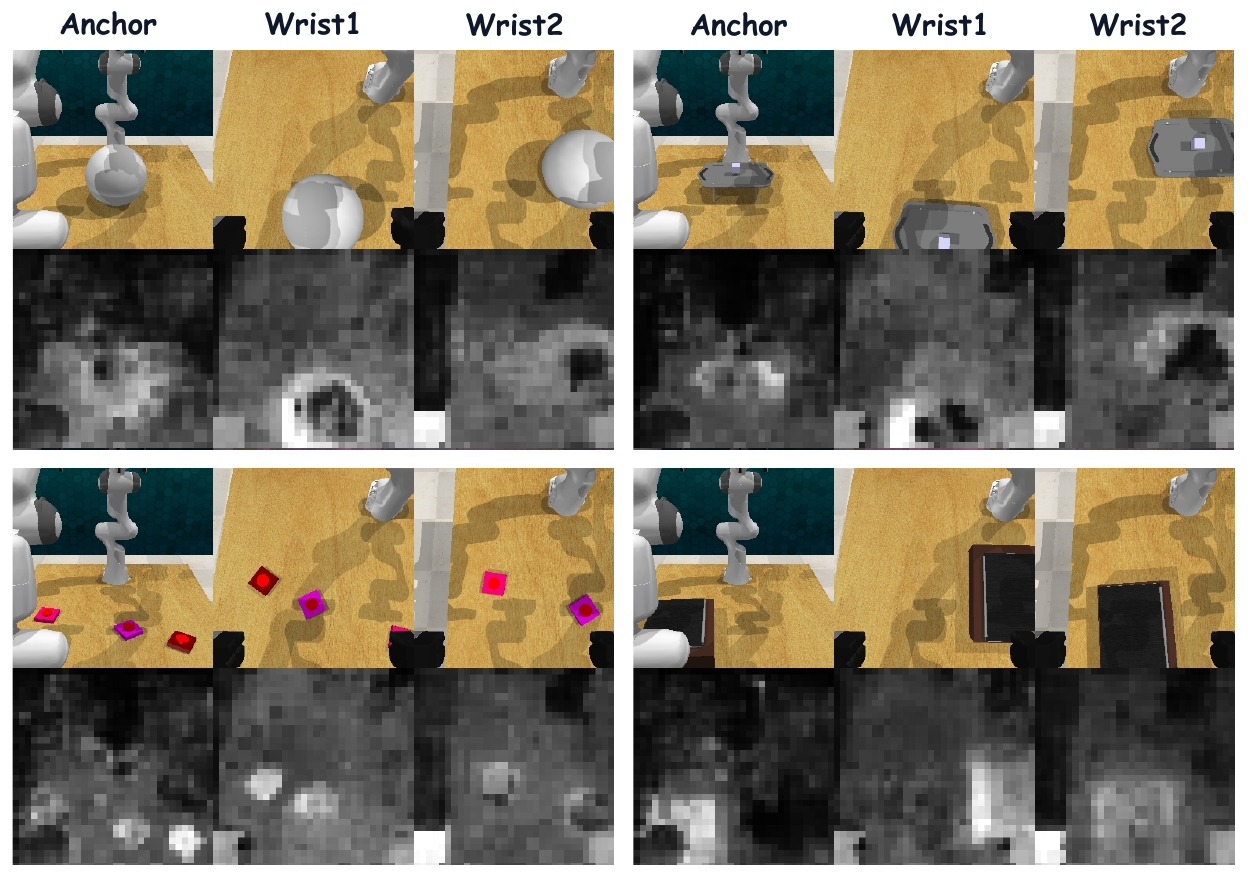}
    \caption{\textbf{Semantic-Spatial Token Interaction visualization.} The figure compares feature responses before and after Semantic-Spatial Token Interaction from Anchor, Wrist1, and Wrist2 views. After attending to spatial tokens, responses concentrate on regions with strong geometric structure, indicating that the spatial stream provides geometric awareness for semantic features.}
    \label{fig:sem2spa_vis}
\end{figure}

Semantic-Spatial Token Interaction mainly improves position prediction. $\mathrm{pos\_acc@0.01}$ rises from 35.8\%, 60.6\%, 81.0\% (baseline) to 42.2\%, 70.0\%, 82.6\% at the three checkpoints, with larger gains than Multi-view Semantic Interaction on all position metrics. Fig.~\ref{fig:sem2spa_vis} shows that the feature responses become more concentrated around geometrically informative regions after semantic-spatial interaction, supporting the improvement in position prediction.

The ``+ Both (no router)'' setting directly merges the semantic and spatial residual streams, which causes interference; semantic and spatial residuals therefore need to be selectively routed according to the prediction head.

With the Head-Aware Routing Gate, the full interaction design achieves the best overall performance. At 50k and 100k it leads on all metrics; at 200k it further reduces $\mathrm{pos\_l2}$ to 0.009, raises $\mathrm{pos\_acc@0.01}$ to 86.3\%, and maintains $\mathrm{rot\_l1}=0.045$ and $\mathrm{gripper}=98.3\%$. This shows that the two interactions are not simply accumulated. Instead, the Head-Aware Routing Gate adaptively routes the residuals for each prediction head: the position head can rely more on the local spatial residual from Semantic-Spatial Token Interaction, while the rotation head can retain the multi-view semantic-consistency residual from Multi-view Semantic Interaction. With this head-conditioned residual routing, semantic and spatial increments jointly optimize the scene tokens and remain complementary across training stages.

\subsubsection{Metric Point-Cloud Construction Ablation}
\begin{figure}[!t]
    \centering
    \includegraphics[width=\linewidth]{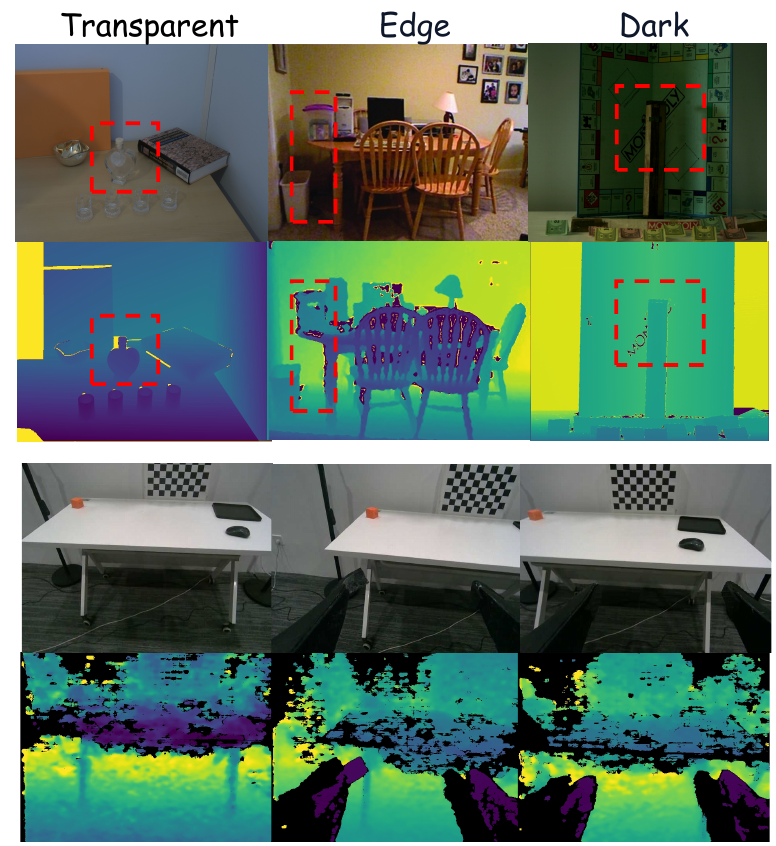}
    \caption{\textbf{Typical depth failures.} Public depth datasets~\cite{booster,nyudepth,middlebury} and depth artifact studies~\cite{ibrahim2020depth} show holes and boundary artifacts caused by transparent or reflective surfaces, low texture, and robot-arm occlusion.}
    \label{fig:depth_failures_ablation}
\end{figure}

MV-Actor back-projects scene tokens into the robot-frame metric space using depth; depth quality therefore directly affects the reliability of all downstream interactions. Fig.~\ref{fig:depth_failures_ablation} summarizes the degradation sources considered in this paper: the upper panels show typical holes and boundary artifacts in public depth datasets, while the lower panels show missing and noisy depth regions observed by the D405 cameras in our real-robot setup. We construct degraded depth by removing depth values with two masks: sparse random holes and contiguous block-missing masks. The block-missing masks simulate large missing regions, while sparse holes simulate small invalid measurements. We report pixel-wise MAE over three regions: \textbf{all} for all pixels, \textbf{miss} for pixels inside the block-missing masks, and \textbf{obs} for valid pixels outside the block-missing masks, which measures whether the repair preserves observed depth outside large holes. Each row in Table~\ref{tab:spatial_ablation_pending} ablates a different combination of guidance inputs: sensor depth alone, adding RGB, adding feed-forward depth (FFD; Pi3X or MoGe2), or combining both. Bold marks the best result in each column.

\begin{table}[!t]
    \centering
    \caption{DEPTH REPAIR ABLATION WITH DIFFERENT GUIDANCE INPUTS. BOLD DENOTES THE BEST RESULT IN EACH COLUMN.}
    \label{tab:spatial_ablation_pending}
    \small
    \setlength{\tabcolsep}{0pt}
    \renewcommand{\arraystretch}{1.08}
    \begin{tabular*}{\columnwidth}{@{\extracolsep{\fill}}lccccc@{}}
        \toprule
        Setting & RGB & FFD & \makecell[c]{MAE\\(all)$\downarrow$} & \makecell[c]{MAE\\(miss.)$\downarrow$} & \makecell[c]{MAE\\(obs.)$\downarrow$} \\
        \midrule
        Sensor only & $\times$ & $\times$ & 0.150 & 1.146 & 0.010 \\
        + RGB & \checkmark & $\times$ & 0.014 & 0.049 & 0.009 \\
        + Pi3X & $\times$ & Pi3X~\cite{pi3} & 0.030 & 0.160 & 0.012 \\
        + Both (Pi3X) & \checkmark & Pi3X~\cite{pi3} & \textbf{0.012} & 0.045 & \textbf{0.008} \\
        + MoGe2 & $\times$ & MoGe2~\cite{moge2} & 0.012 & 0.037 & 0.009 \\
        + Both (MoGe2) & \checkmark & MoGe2~\cite{moge2} & 0.014 & \textbf{0.035} & 0.010 \\
        \bottomrule
    \end{tabular*}
\end{table}

\begin{figure}[!t]
    \centering
    \includegraphics[width=\linewidth]{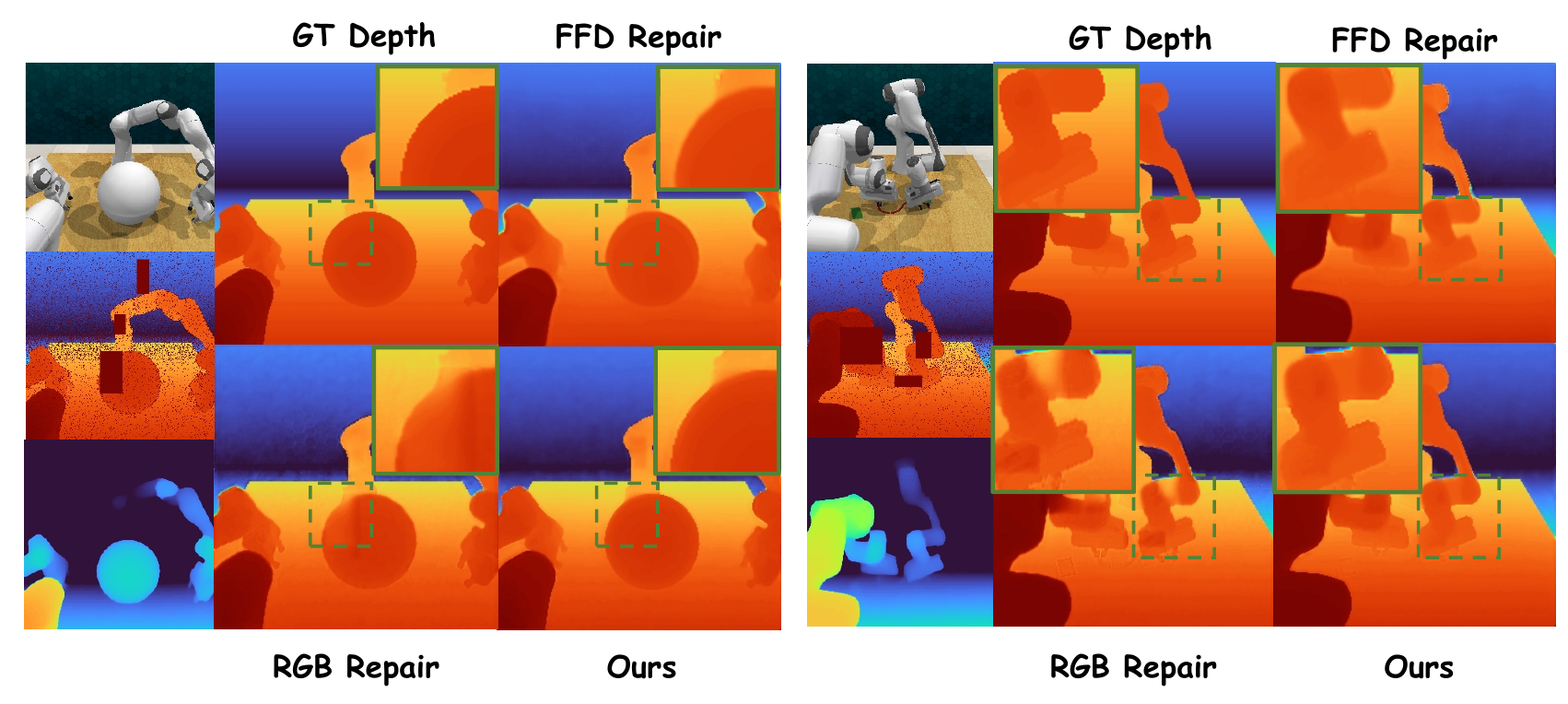}
    \caption{\textbf{Qualitative depth repair.} Each example shows RGB, degraded sensor depth, feed-forward depth (FFD), ground-truth depth, FFD-only repair, RGB-only repair, and RGB+FFD repair. Joint guidance better balances hole filling, boundary recovery, and global structure consistency.}
    \label{fig:depth_est}
\end{figure}

RGB+Pi3X achieves the lowest \textbf{all} and \textbf{obs} error, giving the best overall metric point cloud quality, while RGB+MoGe2 achieves the lowest \textbf{miss} error for large-hole recovery. Both joint-guidance settings substantially outperform single-guidance counterparts because RGB provides local boundary and texture cues, whereas FFD provides a global geometric prior for structurally coherent estimates inside large holes. Qualitative results in Fig.~\ref{fig:depth_est} further visualize how joint guidance preserves object boundaries while recovering continuous depth in missing regions. The module also generalizes across different FFD sources (Pi3X and MoGe2), indicating that the repair design is not tied to a specific feed-forward depth model.

\section{Conclusion}
In this paper, we studied how bimanual manipulation policies can better use the rich information provided by multi-camera observations. We observed that existing policies either treat different views as independent visual inputs or depend on fragile sensor-depth alignment, which limits their ability to build a unified perception of the scene. 

To address this problem, we proposed MV-Actor, a multi-view perception framework that treats different camera views as related observations. MV-Actor unifies calibrated multi-view observations by sharing semantic perception across views and grounding it with reliable spatial awareness from feed-forward reconstruction model.

Experiments on PerAct2 and real-robot tasks show that MV-Actor's semantic-spatial representation improves bimanual manipulation performance over RGB-only and explicit 3D baselines. Our investigation reveals that sharing semantic perception and building reliable spatial awareness across cameras is an effective way to turn multi-view observations into a stronger perceptual basis for bimanual manipulation. Future work will investigate adaptive view selection and temporal consistency to make multi-view semantic-spatial representations more robust under extended manipulation horizons.
\bibliographystyle{unsrt}
\bibliography{references}
\clearpage
\appendix
\begin{center}
\Large\bfseries SUPPLEMENTARY MATERIAL
\end{center}
\vspace{0.4em}
\appsection{app:module_details_en}{Module-Level Implementation Details}

\paragraph{Semantic tokens and anchors}
Frozen CLIP RN50~\cite{resnet} features with an FPN~\cite{fpn} \texttt{res3} readout form a $32\times32$ token grid per camera; repaired depth back-projects each token to a robot-frame 3D anchor.

\paragraph{Cross-view Semantic Interaction}
We use 2 cross-attention layers with heads $=8$. Physical matches are selected by calibrated reprojection with radius $=1$, maximum reprojection error $=1.5$, and maximum depth error $=0.05$ m.

\paragraph{Semantic-spatial interaction}
We use pointwise spatial readout with heads $=8$. Pi3X spatial tokens are projected from 1024 dimensions to $d$, aligned to the $32\times32$ semantic-token locations, and injected through a sigmoid-gated spatial residual with residual scale $=0.1$.

\paragraph{Head-aware routing}
A two-layer SiLU MLP predicts four sigmoid gates, \texttt{sem\_pos}, \texttt{spa\_pos}, \texttt{sem\_rot}, and \texttt{spa\_rot}, to route semantic and spatial residuals to the position and rotation heads.

\appsection{app:impl_en}{Implementation Details}

Unless otherwise specified, all simulation experiments use three calibrated views: front, left wrist, and right wrist. Table~\ref{tab:appendix_policy_hparams_en} lists the policy configuration used by the full MV-Actor model.

\begin{table*}[t]
    \centering
    \caption{Key hyperparameters for MV-Actor. Values correspond to the three-view training configuration used in this paper.}
    \label{tab:appendix_policy_hparams_en}
    \small
    \setlength{\tabcolsep}{6pt}
    \renewcommand{\arraystretch}{1.08}
    \begin{tabular*}{\textwidth}{@{\extracolsep{\fill}}p{0.28\textwidth}p{0.66\textwidth}@{}}
        \toprule
        Item & Setting \\
        \midrule
        Dataset & PerAct2 \\
        Cameras & $V=3$ views: \texttt{front}, \texttt{wrist\_left}, \texttt{wrist\_right} \\
        Input resolution & $256\times256$ RGB and depth \\
        History steps & $H_{\mathrm{obs}}=3$ \\
        Action chunk & Single-step keypose-only action prediction \\
        Temporal subsampling factor & 5 \\
        Token embed dim & $d=120$ \\
        \midrule
        RGB encoder & Frozen CLIP RN50, FPN readout at \texttt{res3} \\
        Text encoder & Frozen CLIP text encoder, instruction tokens pre-tokenized \\
        Semantic token grid & $32\times32$ CLIP-FPN tokens per view \\
        Spatial token source & Frozen Pi3X pointmap/decoder features \\
        Spatial token grid & $18\times18$ Pi3X tokens per view, $d_{\mathrm{spa}}=1024$ \\
        Policy backbone & Denoising Transformer with bimanual action prediction; visual and language tokens are pre-extracted offline \\
        Vision-language attention & 3 cross-attention layers, heads $=8$, FFN width $=4d$ \\
        \midrule
        Cross-view semantic aggregation branch & 2 cross-attention layers, heads $=8$ \\
        Cross-view semantic aggregation matching & Reprojection matching on the $32\times32$ token grid with radius $=1$, max reprojection error $=1.5$ grid cells, max depth error $=0.05$ m \\
        Cross-view semantic aggregation residual & Residual scale $=0.1$ \\
        Semantic-spatial interaction branch & Pointwise spatial readout, heads $=8$ \\
        Semantic-spatial interaction residual & Spatial-awareness residual, residual scale $=0.1$ \\
        Spatial alignment & Pi3X tokens are projected to $d$ and resized from $18\times18$ to $32\times32$ by nearest-neighbor token copying before pointwise readout \\
        Router proprio projection & Bimanual proprio history dimension $=48$, linear projection $48\rightarrow d$ \\
        Router MLP & 2 linear layers: $(5d+6)\rightarrow d\rightarrow4$, hidden activation SiLU \\
        Router output & Separate semantic/spatial gates for position and rotation heads; output bias init $=0$, weight std $=10^{-3}$ \\
        \midrule
        Trajectory-language attention & 1 cross-attention layer, heads $=8$, FFN width $=4d$ \\
        Trajectory-scene attention & 2 cross-attention layers, heads $=8$, FFN width $=d$, with AdaLN \\
        Shared action decoder & 4 self-attention layers, heads $=8$, FFN width $=d$, with AdaLN \\
        Position / rotation heads & 2 self-attention layers per head, MLP predictor $d\rightarrow d\rightarrow3/6$ \\
        Gripper head & MLP predictor $d\rightarrow d\rightarrow1$ \\
        Flow model & Rectified flow, 5 denoising steps \\
        Rotation representation & Quaternion $(x,y,z,w)$ \\
        \midrule
        Optimizer & AdamW, betas $(0.9,0.95)$, weight decay $1\times10^{-10}$ \\
        Learning rate & Main lr $1\times10^{-4}$, backbone lr $1\times10^{-6}$, constant schedule \\
        Total batch size / val batch size & 256 / 64 \\
        Max step & 200k updates for ablations, 400k updates for the full MV-Actor \\
        Training hardware/time & 8 $\times$ A100 GPUs; about 20h for the 400k full-model run \\
        Validation / checkpoint & Validation every 10k updates, intermediate checkpoint every 20k updates \\
        \bottomrule
    \end{tabular*}
\end{table*}

Table~\ref{tab:appendix_depth_repair_hparams_en} gives the depth-repair configuration. The repair model is pre-trained separately and frozen during policy training.

\begin{table*}[t]
    \centering
    \caption{Training settings for Pi3X-guided metric depth repair. The repair network is first pre-trained on PerAct2 simulation data and then frozen during joint policy training.}
    \label{tab:appendix_depth_repair_hparams_en}
    \small
    \setlength{\tabcolsep}{6pt}
    \renewcommand{\arraystretch}{1.12}
    \begin{tabular*}{\textwidth}{@{\extracolsep{\fill}}p{0.26\textwidth}p{0.68\textwidth}@{}}
        \toprule
        Item & Setting \\
        \midrule
        \multicolumn{2}{@{}l}{\textit{Data \& Input}} \\
        Training data & Depth repair data built from PerAct2 simulation \\
        Train/test split & train:test = 90:10 \\
        Inputs & RGB image, degraded sensor depth, Pi3X feed-forward depth \\
        Input mask policy & Supervision weighting and region-wise evaluation only \\
        \midrule
        \multicolumn{2}{@{}l}{\textit{Network Architecture}} \\
        Overall structure & Three-branch U-Net encoder--decoder \\
        Encoder & 5 scales; two $3\times3$ Conv-BN-ReLU blocks per scale \\
        Channel width & Base width 24; scale-wise channels are 24, 48, 48, 96, and 96 \\
        Downsampling & Stride-2 convolution at scales 2--5 \\
        Multi-scale fusion & $1\times1$ reduction + parallel $3\times3$, $5\times5$, and dilation-$2$ $3\times3$ depthwise convolutions \\
        Decoder & 4 bilinear upsampling blocks with skip connections and two $3\times3$ Conv-BN-ReLU blocks \\
        Output head & 2 residual blocks + $3\times3$ convolution + ReLU \\
        \midrule
        \multicolumn{2}{@{}l}{\textit{Training Settings}} \\
        Optimizer & AdamW \\
        Learning rate & $8\times10^{-4}$ \\
        Weight decay & $1\times10^{-4}$ \\
        Epochs & 100 \\
        Batch size & 8 \\
        Random seed & 42 \\
        Loss function & Pixel-wise $L_1$ depth loss with missing-pixel weight $1+4m$ \\
        Full guidance & RGB texture guidance + Pi3X feed-forward depth guidance \\
        \bottomrule
    \end{tabular*}
\end{table*}

\appsection{app:depth_arch_en}{Depth Repair Network Architecture}

The depth repair module is a three-branch U-Net~\cite{unet} that jointly encodes RGB, degraded sensor depth, and Pi3X feed-forward depth. Each branch uses five encoder scales with channel widths 24, 48, 48, 96, and 96. At every scale, the three streams are fused by a multi-scale block using $1\times1$ channel reduction and parallel $3\times3$, $5\times5$, and dilation-$2$ $3\times3$ depthwise convolutions. A four-stage decoder with skip connections restores full resolution, and the output head predicts a single-channel metric depth map supervised by an $L_1$ loss with higher weight on corrupted pixels.

\appsection{app:peract2_tasks_en}{PerAct2 Task Abbreviations and Evaluation Scope}

The main comparison table follows the task abbreviations used by prior PerAct2 and 3DFA comparisons. Table~\ref{tab:appendix_peract2_tasks_en} expands each abbreviation and clarifies the main capability tested by each task. The task-level grouping is not used for metric computation; it is included only to make the comparison table easier to read.

Baseline results in Table~\ref{tab:main_results} are cited from the 3DFA comparison table. Some prior methods did not report every PerAct2 task; those cells are marked with dashes rather than filled by interpolation. Best and second-best marks are computed independently for each task column among reported entries.

\begin{table*}[t]
    \centering
    \caption{PerAct2 task abbreviations used in Table~\ref{tab:main_results}.}
    \label{tab:appendix_peract2_tasks_en}
    \small
    \setlength{\tabcolsep}{5pt}
    \renewcommand{\arraystretch}{1.12}
    \begin{tabular*}{\textwidth}{@{\extracolsep{\fill}}p{0.08\textwidth}p{0.22\textwidth}p{0.62\textwidth}@{}}
        \toprule
        Abbrev. & Task name & Main capability stressed \\
        \midrule
        PB & \textit{push\_box} & Contact-aware pushing and coarse spatial alignment \\
        LB & \textit{lift\_ball} & Coordinated grasping and lifting of a shared object \\
        DB & \textit{push\_buttons} & Two-arm coordination under separated target regions \\
        PP & \textit{pick\_plate} & Stable grasping around thin object boundaries \\
        ID & \textit{put\_item\_into\_drawer} & Object-part localization and drawer interaction \\
        BF & \textit{put\_bottle\_into\_fridge} & Object retrieval under constrained workspace geometry \\
        HO & \textit{handover\_item} & Cross-arm transfer and stable object identity tracking \\
        PL & \textit{pick\_up\_laptop} & Long-horizon manipulation with changing visibility \\
        SR & \textit{straighten\_rope} & Deformable-object geometry and endpoint reasoning \\
        SW & \textit{sweep\_dust} & Long-horizon tool-object interaction \\
        LT & \textit{lift\_tray} & Bimanual support and global object stability \\
        HE & \textit{handover\_item\_easy} & Simplified cross-arm handover \\
        TO & \textit{take\_tray\_out\_of\_oven} & Long-horizon placement under occlusion and workspace constraints \\
        \bottomrule
    \end{tabular*}
\end{table*}

\appsection{app:vla_baselines_en}{VLA Baseline and Deployment Notes}

All real-robot methods are evaluated on the ARK-Lift bimanual robot with the same three RealSense D405 views, workspace, 100 training demonstrations, and 10 test trials per task. MV-Actor uses $256\times256$ calibrated RGB-D observations with repaired depth, while the $\pi_0$ baseline uses RGB-only head/left-wrist/right-wrist images resized to $224\times224$ and a 14-D bimanual end-effector state. This setup keeps task definitions and data budget fixed while varying only the policy representation.

\appsection{app:pi0_prompts_en}{Language Prompts for $\pi_0$}

Language-conditioned policies use task-level natural-language commands rather than stage-by-stage scripts. In PerAct2 simulation, each episode is paired with a short instruction describing the target bimanual task, following the benchmark protocol. For the real-robot $\pi_0$ baseline, we use one fixed imperative instruction for each task to keep language conditioning consistent across all evaluation trials. The prompts used in our real-robot experiments are listed in Table~\ref{tab:appendix_pi0_prompts_en}.

\begin{table*}[!t]
    \centering
    \caption{Language instructions used by the real-robot $\pi_0$ baseline.}
    \label{tab:appendix_pi0_prompts_en}
    \small
    \setlength{\tabcolsep}{5pt}
    \renewcommand{\arraystretch}{1.12}
    \begin{tabular*}{\textwidth}{@{\extracolsep{\fill}}p{0.16\textwidth}p{0.20\textwidth}p{0.38\textwidth}p{0.18\textwidth}@{}}
        \toprule
        Skill type & Task & Prompt for $\pi_0$ & Note \\
        \midrule
        Cooperative lifting & \textit{lift\_plate} & \emph{lift the plate} & Symmetric bimanual grasp and lift \\
        Contact handover & \textit{handover\_block} & \emph{hand over the cube from the right hand to the left hand} & Cross-arm handover with object identity preservation \\
        Fine spatial alignment & \textit{insert\_stick} & \emph{grasp the stick and insert it into the test-tube rack} & Local geometry alignment and insertion precision \\
        \bottomrule
    \end{tabular*}
\end{table*}

The prompts are intentionally short and task-level. They specify the intended manipulation goal without decomposing the task into intermediate stages, so that the real-robot comparison evaluates the policy and perception stack rather than hand-written procedural guidance. This is especially important for handover and insertion, where explicit stage instructions could otherwise encode additional task structure not available to the other baselines.

\appsection{app:real_robot_protocol_en}{Real-Robot Protocol and Staged Scoring}

Real-robot evaluation is scored by task stages. For every trial, we keep synchronized video, stage screenshots, and a structured score record. Two-stage tasks use equal weights of 0.50 and 0.50. Three-stage tasks use cumulative milestone scores of 0.33, 0.66, and 1.00. Final real-robot results are averaged over 10 independent trials.

\begin{table*}[!t]
    \centering
    \caption{Real-robot task setup and staged scoring protocol. Rows follow the example order shown in Fig.~\ref{fig:appendix_real_robot_snapshots_en}; all tasks use 100 training demonstrations and 10 test trials.}
    \label{tab:appendix_stage_scoring_en}
    \scriptsize
    \setlength{\tabcolsep}{4pt}
    \renewcommand{\arraystretch}{1.08}
    \begin{tabular*}{\textwidth}{@{\extracolsep{\fill}}p{0.15\textwidth}p{0.19\textwidth}cp{0.52\textwidth}@{}}
        \toprule
        Task & Evaluation focus & Stages & Cumulative scoring milestones \\
        \midrule
        \textit{handover\_block} & Contact handover & 2 & S1: the right gripper picks up the target block (0.50); S2: the left and right grippers complete the block handover without dropping it (1.00). \\
        \midrule
        \textit{insert\_stick} & Fine spatial alignment & 3 & S1: the left gripper secures the test tube rack (0.33); S2: the right gripper picks up the stick (0.66); S3: the right gripper inserts the stick into the test tube rack (1.00). \\
        \midrule
        \textit{lift\_plate} & Cooperative lifting & 3 & S1: one gripper grasps the plate (0.33); S2: both grippers grasp the plate (0.66); S3: both grippers lift the plate together (1.00). \\
        \bottomrule
    \end{tabular*}
\end{table*}

For a method $m$ and task $\tau$, let $r_{m,\tau,n}\in[0,1]$ be the cumulative staged score of trial $n$. The percentage score reported in Table~\ref{tab:real_world_results} is
\begin{equation}
S_{m,\tau}=100\cdot\frac{1}{10}\sum_{n=1}^{10} r_{m,\tau,n}.
\end{equation}
The average score is the arithmetic mean over the three real-robot tasks:
\begin{equation}
S_m^{\mathrm{avg}}=\frac{1}{3}\sum_{\tau\in\mathcal{T}_{\mathrm{real}}} S_{m,\tau}.
\end{equation}
This staged protocol gives partial credit for meaningful progress while still reserving the full score for task completion. It is especially useful for bimanual tasks, where a method may reliably complete the first arm-object contact but fail during cross-arm transfer or final insertion.

The real-robot benchmark contains three tasks with the same data budget: 100 training demonstrations and 10 test trials per task. The tasks cover cooperative lifting (\textit{lift\_plate}), contact handover (\textit{handover\_block}), and fine spatial alignment (\textit{insert\_stick}), respectively stressing global support structure, cross-view target identity during transfer, and local insertion precision.

\begin{figure*}[!t]
    \centering
    \includegraphics[width=0.98\textwidth]{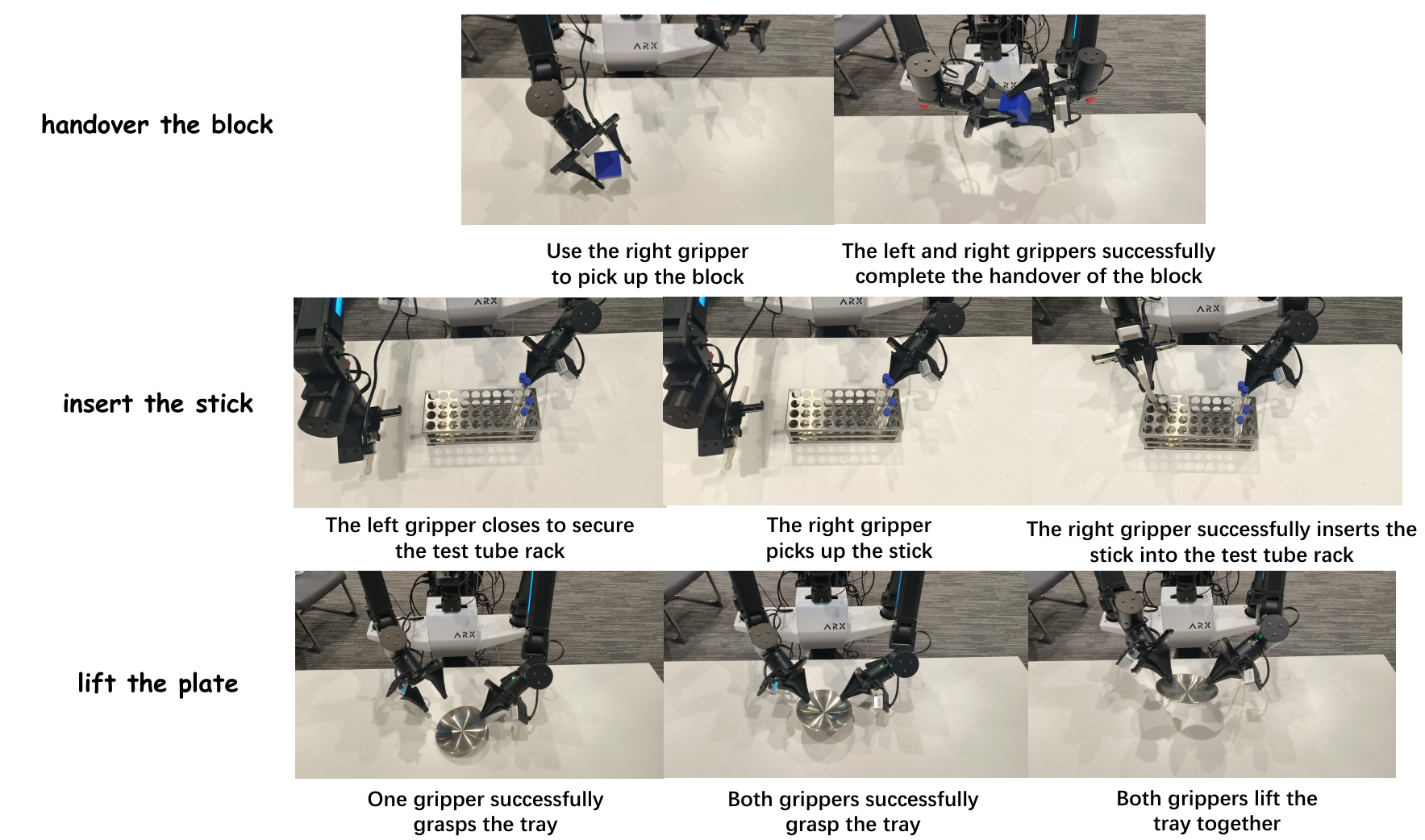}
    \caption{Examples of real-robot evaluation stages, ordered consistently with Table~\ref{tab:appendix_stage_scoring_en}. The rows show handover, stick insertion, and cooperative lifting; the ``tray'' label in the lifting row refers to the plate object used in the \textit{lift\_plate} task.}
    \label{fig:appendix_real_robot_snapshots_en}
\end{figure*}

Table~\ref{tab:appendix_stage_scoring_en} defines the milestones used during evaluation and follows the task order visualized in Fig.~\ref{fig:appendix_real_robot_snapshots_en}: handover, insertion, and lifting. Two-stage handover uses equal stage weights because the main progress points are grasping and transferring. Lifting and insertion use three milestones to reflect contact formation, intermediate stabilization or alignment, and final task completion.

\end{document}